# The Probability of a Possibility: Adding Uncertainty to Default Rules

Craig Boutilier
Department of Computer Science
University of British Columbia
Vancouver, British Columbia
CANADA, V6T 1Z2
email: cebly@cs.ubc.ca

## Abstract

We present a semantics for adding uncertainty to conditional logics for default reasoning and belief revision. We are able to treat conditional sentences as statements of conditional probability, and express rules for revision such as "If $A$ were believed, then $B$ would be believed to degree $p$." This method of revision extends conditionalization by allowing meaningful revision by sentences whose probability is zero. This is achieved through the use of *counterfactual probabilities*. Thus, our system accounts for the best properties of qualitative methods of update (in particular, the AGM theory of revision) and probabilistic methods. We also show how our system can be viewed as a unification of probability theory and possibility theory, highlighting their orthogonality and providing a means for expressing the probability of a possibility. We also demonstrate the connection to Lewis's method of imaging.

## 1 Introduction

Recently, a number of (more or less equivalent) conditional theories for default reasoning and belief revision have been proposed (Kraus, Lehmann and Magidor 1990; Goldszmidt and Pearl 1991; Boutilier 1990, 1992a, 1993a, 1993b). The cornerstone of such conditional logics is a conditional connective $\Rightarrow$. The sentence $A \Rightarrow B$ can be interpreted as a default rule of the form "If $A$ then normally $B$." $A \Rightarrow B$ may also be interpreted as a subjunctive conditional of the form "If an agent were to believe $A$ then it would believe $B$." According to the *Ramsey test* such a conditional is true iff the agent comes to accept $B$ when it revises its beliefs to incorporate a new sentence $A$ (Stalnaker 1968). Thus, any conditional logic can be thought of as determining a theory of belief revision.

Unfortunately, such conditionals cannot be used to represent uncertainty in the revision process. In this paper, we present a system in which one can express conditionals of this form, but attach to the consequent of the conditional a degree of belief to capture the uncertainty inherent in the conclusion. Thus, we will be able to capture statements of the form "If an agent believed $A$, it would believe $B$ to degree $p$." The notion of revision determined by such statements will extend the usual process of conditionalization through the use of counterfactual probabilities.

### 1.1 Conditional Logic and Degrees of Belief

The logics cited above, should we take $\Rightarrow$ to be a subjunctive interpreted according to the Ramsey test, all correspond to the well-known *AGM theory* of belief revision (Alchourrón, Gärdenfors and Makinson 1985; Gärdenfors 1988). In fact, it has been shown that default rules of the above form can be interpreted as subjunctive conditionals (Boutilier 1992c, 1993b; Makinson and Gärdenfors 1990). One need only view default rules as subjunctive conditionals that express constraints on how an agent revises a theory of *expectations*. In what follows, we usually take $A \Rightarrow B$ to be a subjunctive conditional, assuming that such a conditional may be interpreted as a default rule with no difficulty.

A feature of conditional logics that has lead to their success in the representation of defeasible inference is the fact that conditionals can be used to express default rules and subjunctives in a very natural way and can be used to derive new rules. In particular, it is consistent to assert together the conditionals $A \Rightarrow C$ and $A \wedge B \Rightarrow \neg C$, demonstrating the defeasibility inherent in conditional reasoning. This fact can be exploited in the representation of uncertain or context-dependent inference: in context $A$, $C$ is a reasonable conclusion; but if $B$ is known as well, $C$ is no longer acceptable. For example, in certain approaches to model-based diagnosis, preference is given to diagnoses requiring as few faulty components as possible. So a conditional default rule might suggest that, given observation $O_1$, component $C_1$ alone is faulty. Given an additional observation $O_2$ however, $C_1$ might fail to be a diagnosis.

As mentioned above, a difficulty with the conditional approach to default inference and belief revision is its categorical nature. One can represent the fact that $C$ is (fully) believed or that it is not; but no other distinctions can be made, for example, with respect to *degree* of belief. If neither $C$ nor $\neg C$ is fully believed, both are "serious possibilities," but neither can be preferred to the other; so, for example, an agent cannot decide whether to act on the ba-sis of $C$ or $\neg C$. Similar remarks apply to conditionals: if an agent accepts both $\neg(A \Rightarrow C)$ and $\neg(A \Rightarrow \neg C)$ then, should the agent adopt $A$, the relative likelihood of $C$ and $\neg C$ cannot be represented. In our informal example, we might have a default that exactly one of components $C_1$, $C_2$ or $C_3$ is faulty given $O_1$. Thus, in context $O_1$, all multiple faults (and single faults involving other components) are rejected as serious possibilities. This reduces the space of candidate diagnoses, but does not allow one to distinguish the remaining candidates. Typically, we want to investigate alternatives according to their likelihood (de Kleer 1991; Poole 1993). If failure of $C_1$ is more likely than that of $C_2$ or $C_3$, testing strategies (say) might be altered.

Probabilistic representations can be used to circumvent this difficulty. One can assign probabilities to each possibility admitted by a belief set or knowledge base *KB*. We think of such probabilities as *degrees of belief*. A *serious possibility* relative to any belief set is any sentence $A$ such that $P(A) > 0$. A (full) *belief* is any sentence $A$ such that $P(A) = 1$.

With respect to a given *KB*, the conditional $A \Rightarrow C$ means that revising *KB* to accommodate $A$ results in a belief in $C$. Assuming that we have a probability function $P$ assigning degrees of belief to the possibilities admitted by *KB*, it is natural to assume that revising *KB* by $A$ is identical to conditionalization of $P$ by $A$; thus, asserting $A \Rightarrow C$ amounts to stating that $P(C|A) = 1$. Furthermore, it provides a natural means of specifying "uncertain" conditionals: if $P(C|A) = p$ then we might say that a conditional $A \Rightarrow C$ *holds to degree* $p$ (if an agent came to believe $A$ it would believe $C$ to degree $p$). Unfortunately, there is a crucial difficulty with such an approach. If $P(A) = 0$, then $P(C|A)$ is undefined or given some trivial value for all $C$. In contrast, conditionals $A \Rightarrow C$ are meaningful in the case where $\neg A \in \mathit{KB}$. Theories of revision provide nontrivial results, new beliefs, possibilities and impossibilities, even when the new information $A$ is inconsistent with *KB* (indeed, this is the principle case for belief revision). For instance, in model-based diagnosis, typically one *only* looks for a diagnosis when observation $O$ is *inconsistent* with *KB* (typically a system description together with normality assumptions). Conditioning on $O$ is then meaningless.

In order to augment arbitrary conditionals with degrees of belief we need some generalization of conditional probability, whereby it is meaningful to assert $P(C|A) = p$ (for some non-extreme value $p$) when $P(A) = 0$. In this paper we present a semantics for just such a system. We use the notion of *counterfactual probability* as described by Stalnaker (1970) and Lewis (1976). Our system can be viewed as a refinement and a semantic model for counterfactual probabilities. We demonstrate the relationship to existing conditional logics for default reasoning and belief revision. In particular, we show that our logic (conservatively) extends and unifies two predominant views of belief updating: probabilistic conditionalization and the AGM theory of revision. We also describe our system in terms of *possibilistic logic* (Dubois and Prade 1988). As pointed out in (Boutilier 1992b), possibilistic logic cannot be used to represent uncertainty or degrees of belief. Rather it should be viewed as a representation mechanism for the *entrenchment* of certain beliefs (see Section 2); thus it provides a means for updating by counterfactual sentences. In this sense, our logic is a unification of probability theory and possibility theory that highlights their orthogonality and adds to possibility theory the means to express the probability of a possibility.

### 1.2 Why Use Categorical Rules?

If one is going to use probabilities as degrees of belief, it seems natural to question the need for (categorical) default rules, conditionals or counterfactual probabilities. If one is going to allow a sentence $A$ in *KB* to be retracted when $\neg A$ is learned, why not simply assign $A$ some degree of belief less than 1 in the first place and use standard techniques such as conditionalization to incorporate new items of belief?

If one wishes to allow the possibility that any "belief" can be overturned given the proper evidence, then full belief can be granted only to tautologies, and every contingency must have some probability. To take a slightly less extreme view, one might accord observational reports (say) the status of full belief, but still no conclusions drawn from these would be certain. Presumably, there are certain computational advantages to be gained by ruling out possibilities that are very unlikely (Cheeseman 1985; Harman 1986). Chief among these is the ability to exploit logical rules of inference. Such rules allow conclusions to be reached in manner that is independent of context, in contrast to probabilistic inference. The *locality* of logical rules can be exploited if parts of the knowledge base (are assumed to) have full belief (Pearl 1988).

It may also be that the cost associated with reaching incorrect (unhedged) conclusions and being forced to revise the belief set is outweighed by the probability of being correct. We might therefore think of a default rule as an instantiation of an *acceptance rule* (Kyburg 1961). If $A \Rightarrow B$ is a conditional held by the agent, we take it to mean that there is a certain utility associated with complete acceptance of $B$ given $A$.[1] On this view, it is reasonable to allow a conditional $A \Rightarrow B$ to be held even when $\neg A$ is accorded full belief, $P(\neg A) = 1$. Consequently, we do not take a $P(A) = 0$ to indicate that $A$ is (logically or physically) impossible, but simply that is is not, to use Levi's (1980) terminology, a *serious possibility*.

### 1.3 Overview

In Section 2 we review the possible worlds semantics for epistemic states, conditional logics and belief revision, showing its strong relation to possibility theory. In Section 3 we add probabilities to this system in such a way that the conditional probability $P(B|A)$ is meaningful even when $P(A) = 0$. We show that our method of belief revision extends the AGM theory so that degrees of belief can be represented in revised belief sets, and that revision is identical to conditionalization if the update $A$ is a serious possibility (i.e., if $P(A) > 0$). We also demonstrate how one may "index" degrees of possibility so that correct inferences can be simulated without resorting to explicit counterfactual probabilities (thus, standard probabilistic reasoning techniques can be applied to this reasoning process). In Section 4 we illustrate the relationship of our system to the method of *imaging* proposed by Lewis (1976). In particular, we show that our revision method fits the pattern of both generalized imaging and conditionalization. This stands in contrast to a widely-held view that conditionalization and generalized imaging are irreconcilable (Gärdenfors 1988). We conclude with a discussion of updated counterfactual probabilities and iterated revision.

---

[1] We do not address here the issue of how one determines appropriate acceptance rules; but in general decision-theoretic criteria should be brought to bear.

## 2  Belief Revision and Possibilistic Logic

To keep the presentation simple, we assume a propositional language $\mathbf{L}_{CPL}$ generated by a finite set of atomic variables **P**. The set of classical valuations for this language is denoted $V$, elements of $V$ usually referred to as *worlds*. We take an agent to possess a deductively closed set of beliefs $K$ (typically the closure of some finite knowledge base *KB*). The usual semantics for belief models the agent's belief state as a set of *epistemically possible worlds*, those worlds that make each belief in $K$ true. Using the modal connective B for belief, an *epistemic state* (set of worlds) $W$ satisfies $BA$ just when $A$ is true at each world in $W$.[2]

We introduce several definitions. The set of $A$-worlds is denoted $\|A\| = \{w \in V : w \models A\}$. For a set of formulae $K$, the set of $K$-worlds $\|K\|$ is simply the set of worlds satisfying each element of $K$.

### 2.1  Belief Revision

This model of epistemic states is reasonable as long as an agent's beliefs never strain the credibility of $K$. However, should an agent learn some sentence $A$ such that $\neg A \in K$, some revision of the agent's belief set and epistemic state is required. One of the best known theories of revision of this type is the *AGM theory* (Alchourrón, Gärdenfors and Makinson 1985; Gärdenfors 1988). In its most widely cited form, the theory is presented as a set of postulates constraining logically acceptable revision functions. If $K$ is some belief set, $K_A^*$ denotes the belief set resulting from the revision of $K$ to include $A$. One of the hallmarks of the AGM theory is its commitment to the principle of *minimal change*: one should give up as few beliefs as possible in $K$ when attempting to accommodate $A$. The postulates impose certain logical constraints on this notion. A key and characteristic property of AGM revision operators is that $K \subseteq K_A^*$ whenever $\neg A \notin K$: if no beliefs have to be given up in order to accept $A$, then none should be. We call any revision function $*$ that satisfies the postulates and *AGM revision operator*.

In (Boutilier 1992a, 1993b) we present a possible worlds semantics and modal logic for the representation of AGM revision functions. A *CO-model* consists of a set of worlds $W \subseteq V$ and an ordering relation $\preceq$ on $W$ reflecting the degree of plausibility attributed to various worlds by an agent. We interpret $w \preceq v$ as "$w$ is at least as plausible as $v$." We insist that $\preceq$ be a total preorder on $W$ (that is, a transitive, connected binary relation — see below). Intuitively, the most plausible worlds in $W$ are those consistent with the agent's beliefs. All other worlds are epistemically impossible, but some are more plausible than others. A CO-model is a $K$-*revision model* just in case the set of most plausible worlds, those minimal in $\preceq$, is exactly $\|K\|$. It is just these models that can be reasonably used to represent and belief set $K$ and its revision.

Should an agent learn some information that contradicts its beliefs, $\|K\|$ can no longer be held as a reasonable epistemic state. If $\neg A \in K$ and $A$ is learned, the agent must adopt a new belief set $K_A^*$ of which $A$ is a member. Semantically, we simply require that the agent's new epistemic state be represented by the set of *most plausible $A$-worlds*, those minimal in $\preceq$. Should $B$ be true at each such world we say that $B \in K_A^*$. According to the Ramsey test, this captures the acceptance conditions for a subjunctive conditional $A \Rightarrow B$: if the agent believed $A$, it would believe $B$. In (Boutilier 1992a, 1993b) we provide a strong representation theorem relating this model to the AGM theory. We provide further technical details below.

### 2.2  Possibilistic Logic

In (Boutilier 1992b) we show how the modal logic CO can be used to capture qualitative possibilistic logic. We review this connection here, and formalize the model of revision discussed above in possibilistic terms.

Possibilistic logic has been developed to a great extent by Dubois and Prade (see their (1988) for a survey). A *possibility measure* $\Pi$ maps the sentences of $\mathbf{L}_{CPL}$ into the real interval $[0, 1]$. The value $\Pi(\alpha)$ is intended to represent the degree of possibility of $\alpha$. We take this to represent the amount of surprise associated with adopting $\alpha$ as an epistemic possibility. If $\Pi(\alpha) = 1$ there is no surprise (i.e., $\alpha$ is consistent with the agent's beliefs), while $\Pi(\alpha) = 0$ indicates that surprise is maximal (i.e., an agent would *never* adopt $\alpha$). A possibility measure must satisfy the following three properties:

(a)  $\Pi(\top) = 1$

(b)  $\Pi(\bot) = 0$

(c)  $\Pi(A \vee B) = \max(\Pi(A), \Pi(B))$

A *necessity measure* $N$ is a similar mapping, associating with $\alpha$ a degree of necessity. We take $N(\alpha)$ to represent the amount of surprise associated with giving up belief in $\alpha$ (or the degree of *entrenchment* of $\alpha$ in a belief set; see (Boutilier 1992b)). One may define necessity measures using the identity

$$N(\alpha) = 1 - \Pi(\neg \alpha).$$

Semantically, we can model possibility measures using *possibility distributions*. A distribution $\pi$ assigns to each world in $V$ a degree of possibility from the interval $[0, 1]$. This can be viewed as a ranking of worlds, with $w$ being at least as possible as $v$ just when $\pi(w) \geq \pi(v)$. This corresponds precisely to the relation $w \preceq v$ in a CO-model.

---

[2] A somewhat more involved notion of satisfaction is required to capture iterated belief sentences, but this will suffice for our informal presentation.

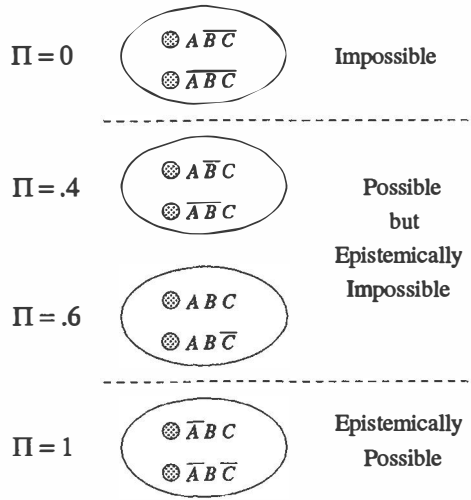

Figure 1: A Possibility Model

**Definition 1** A *possibility model* is a triple $M = \langle V, W, \pi \rangle$ where

(a) $V$ is the set of worlds suitable for $\mathbf{L}_{CPL}$;
(b) $\pi$ maps $V$ into $[0, 1]$; and
(c) $W = \{w \in V : \pi(w) > 0\}$.

Figure 1 illustrates a possibility model. $W$ is the set of possible worlds, those assigned a non-zero degree of possibility. Epistemically possible worlds are those assigned possibility 1. We can also define truth conditions for a belief operator B and a subjunctive conditional $\Rightarrow$.

**Definition 2** Let $M$ be a possibility model.
(a) World $w$ is *epistemically possible* iff $\pi(w) = 1$.
(b) $M$ induces the *belief set* $K$, where $K$ is characterized by $\|K\| = \{w \in W : \pi(w) = 1\}$.
(c) $M \models \mathsf{B}A$ iff $A \in K$
(iff $\{w \in W : \pi(w) = 1\} \subseteq \|A\|$).
(d) The set of *plausible* (or most possible) $A$-worlds is denoted $Pl(A)$, where $w \in Pl(A)$ iff $w \models A$ and $\pi(w) \geq \pi(v)$ for all $v \models A$.
(e) $M \models A \Rightarrow B$ iff $Pl(A) \subseteq \|B\|$.
(f) $K_A^* = \{B : M \models A \Rightarrow B\}$.

The model in Figure 1 captures the belief set $K = Cn(\{\neg A, B\})$. The conditional $A \Rightarrow B$ is true, while neither $A \Rightarrow C$ nor $A \Rightarrow \neg C$ hold. This is because $B$ holds at all of the most plausible $A$-worlds (those with possibility .6), while $C$ and $\neg C$ do not. We can show that the class of revision function $*$ induced by possibility models is exactly the class of AGM revision functions.[3]

---
[3]Technically, we require that the possibility model be "complete" in the sense that $V = W$, that is, all worlds are possible.

**Theorem 1 (Boutilier 1992b)** *For any possibility model $M$, the induced revision function $*$ satisfies the AGM postulates. For any AGM revision operator $*$ and belief set $K$, there is a possibility model $M$ that induces $*$.*

A distribution determines a possibility measure $\Pi$ as follows:
$$\Pi(A) = \max\{\pi(w) : w \models A\}.$$
In other words, the degree of possibility of $A$ is just that of the *most* possible $A$-worlds (i.e., those worlds in $Pl(A)$). The model in Figure 1 admits $\neg A$, $B$, $C$ and $\neg C$ as serious possibilities (e.g., $\Pi(C) = 1$), while $A$ and $\neg B$ are not ($\Pi(A) = .6$ and $\Pi(\neg B) = .4$). Notice that $\neg B \wedge \neg C$ is *impossible*: $\Pi(\neg B \wedge \neg C) = 0$. Assuming that $\pi(w) = 1$ for some $w \in V$,[4] $\Pi$ determines a consistent belief set $K$. Furthermore, we have the following obvious relationships:

(a) $A \in K$ iff $N(A) > 0$ iff $\Pi(\neg A) < 1$. Such a sentence is said to be *accepted*. If $\neg A \in K$, $A$ is said to be *rejected*.
(b) $A \notin K$ iff $\Pi(\neg A) = 1$. If neither of $A$ or $\neg A$ is in $K$, $A$ is said to be *indeterminate*.
(c) $M \models A \Rightarrow B$ iff $\Pi(A \wedge B) > \Pi(A \wedge \neg B)$ or $\Pi(A) = 0$.

We note that possibility rankings can be given a probabilistic interpretation using $\varepsilon$-semantics (Adams 1975; Pearl 1988). We can associate $\kappa$-rankings (Goldszmidt and Pearl 1991) with degrees of possibility and take $\pi(w) > \pi(v)$ to mean that world $w$ is arbitrarily more probable than $v$. In this manner, we can ensure that the conditional probability $P(B|A)$ can be made arbitrarily high if $A \Rightarrow B$ holds (Adams 1975).

## 3 Counterfactual Probabilities

The notion of epistemic state introduced in the last section has certain drawbacks. In particular, the serious possibilities held by an agent cannot be distinguished according to their degree of certainty or belief. The epistemic status of a proposition $A$ is one of acceptance, rejection or indeterminacy. Indeterminate propositions reflect an uncertainty about their truth, but among indeterminate propositions, no distinctions can be made with respect to *degree* of uncertainty. Notice that possibility theory has nothing to offer in this regard: if neither $A$ nor $\neg A$ are believed then both have degree of possibility 1 and degree of necessity 0.

### 3.1 Semantics

One would like to be able to express, for a given belief set $K$, the degree of belief associated with uncertain possibilities. While neither of $A$ or $\neg A$ may be sufficiently likely to warrant full acceptance, evidence may render one more

---
If we drop that restriction, a simple modification of the AGM postulates will suffice.

[4]Throughout, we will assume that $W \subseteq V$ is nonempty. These *proper* possibility models will correspond to nonempty (consistent) epistemic states.

probable than the other. We would like to say, for example, that an agent associates probability .75 with $A$ (or that $A$ is believed to degree .75). Semantically, this is easily accomplished. In the usual fashion, we can assign a non-zero probability weight to each epistemically possible world, taking epistemically impossible worlds ($\pi(w) < 1$) to have probability 0. In this manner, all full beliefs (accepted sentences) have probability 1 and only serious possibilities have non-zero probability.

While this provides degrees of belief for epistemic possibilities, it does nothing to determine counterfactual probabilities (e.g., $P(B|A)$ where $\Pi(A) < 1$). While the possibility distribution $\pi$ distinguishes worlds according to their degree of possibility, worlds within each possibility rank are indistinguishable. However, we can apply the same idea and assign, within each possibility rank, relative weights to worlds. Of course, we need not use a different weight assignment function for each value in the range of $\pi$. We can simply assign weight to all worlds and compare only the weights of worlds with the same degree of possibility.

**Definition 3** A *counterfactual probability model* (CPM) has the form $M = \langle V, W, P, \pi \rangle$ where

(a) $\langle V, W, \pi \rangle$ is a possibility model; and

(b) $P$ maps $W$ into $(0, 1]$

As before, $\pi(w)$ is the degree of possibility assigned to world $w \in V$ and $W$ is the subset of "possible" worlds in $V$. $P(w)$ is the *probability weight* assigned to possible world $w$. This weight must be non-negative, for we assume that only impossible worlds $(V - W)$ can have no weight. The definition of the categorical belief set $K$ and the truth conditions for the conditional connective $\Rightarrow$ are exactly as for a possibility model (Definition 2). The definition of a revised belief set $K_A^*$ also remains unchanged.

As stated earlier, an agent's epistemic state is captured by those worlds $w$ such that $\pi(w) = 1$. Furthermore, should an agent come to accept $A$, its revised belief state is captured by the set of most possible $A$-worlds $Pl(A)$. Given this belief state, the degree of belief accorded some sentence $B$ ought to be the relative weight of all $B$-worlds in this set. This leads to the following definitions.

**Definition 4** Let $M = \langle V, W, P, \pi \rangle$ be a CPM. The *counterfactual probability* of $B$ given $A$ (w.r.t. $M$) is

$$P(B \uparrow A) = \frac{\sum \{P(w) : w \in Pl(A) \text{ and } w \models B\}}{\sum \{P(w) : w \in Pl(A)\}}$$

Since $P$ is defined only on worlds in $W$, this term is undefined iff $A$ is impossible (i.e., if $\pi(A) = 0$).

**Definition 5** The *factual probability* of $A$ (w.r.t. $M$) is

$$P(A) = P(A \uparrow \top)$$

**Proposition 2** *For any CPM $M$, the factual probability function $P$ is a probability function.*

We take the unconditional, factual probability function $P$ to define the *objective* epistemic state of the agent. This is the usual notion of degree of belief, where only epistemic or serious possibilities have non-zero probability.

**Proposition 3** $P(A) = 1$ *iff* $A \in K$.

We define factual conditional probability in the usual way:

**Definition 6** $P(B|A) = \frac{P(A \wedge B)}{P(A)}$ for all $A$ such that $P(A) > 0$.

Our goal is now to describe a method for an agent to move from one epistemic state to another during the course of revision. In particular, we must describe the new factual probability function $P_A^*$ that results when the agent's original epistemic state $P$ is revised by $A$. This revision will proceed by means of *counterfactual conditionalization*.

**Definition 7** Let $P$ be the factual probability function determined by $M$. The *revised factual probability function* $P_A^*$ is given by

$$P_A^*(B) = P(B \uparrow A)$$

The (objective) epistemic state of an agent after such a revision is characterized by $P_A^*$. We can show the following key results. First, $P_A^*$ determines a valid epistemic state:

**Proposition 4** *If $\Pi(A) > 0$ then $P_A^*$ is a probability function.*

We also have that the revised probability function respects the truth conditions for our conditional connective, and hence corresponds precisely to the belief set $K_A^*$:

**Theorem 5** $M \models A \Rightarrow B$ *(i.e., $B \in K_A^*$) iff $P_A^*(B) = 1$.*

Given the representation result Theorem 1, we can show that this model forms a proper extension of the AGM theory of revision. For any CPM $M$, let the $*$ denote the revision function induced by the $M$ (i.e., $K_A^* = \{B : P_A^*(B) = 1\}$).

**Corollary 6** *The induced revision function $*$ satisfies the AGM postulates; and for any $K$ and AGM revision function $*$, there is a CPM that induces $*$.*

Finally, it's not hard to see that $P_A^*$ respects the usual notion of conditionalization when this is applicable:

**Theorem 7** *If $P(A) > 0$ then $P_A^*(B) = P(B|A)$.*

Thus CPMs can be viewed in several different ways. First, they extend the AGM theory of revision (and hence conditional theories of default reasoning) with the power to express uncertain conclusions probabilistically. Second, they can be thought of as a means of representing counterfactual probabilities and extending the notion of conditionalization to conditions with zero probability. Third, they unify possibility and probability theory in a way that highlights the orthogonal roles they have to play in the representation of uncertain information and inference.

Notice that while we have specified here an updated *factual* probability function, we have said nothing about the new *counterfactual* probabilities an agent should adopt upon learning $A$. There are several ways in which one might proceed. We discuss this issue in the concluding section.

### 3.2 Using Factual Probabilities

Given this semantics for counterfactual probabilities, the question remains: how should one compute the result of updating a belief set? Given that one has standard techniques for dealing with probability measures, these can be used in (at least) two ways to *simulate* this model of counterfactual probabilities. The first fairly obvious method is to assign a unique probability function $P_k$ for each degree of possibility in the range of $\pi$ (i.e., for each $k = \pi(w)$ for some $w$).[5] The functions $P_k$ in this sequence need only satisfy the property that no two distinct functions assign positive probability to the same maximal conjunction of literals. In other words, if $P_k(w) > 0$ then $P_j(w) = 0$ for all $j \neq k$. Call such a sequence of functions *admissible*. The *most possible function* for sentence $A$ in this sequence, denoted $P_A$, is the function $P_k$ where

$$k = \max\{i : P_i(A) > 0\}$$

When revising by $A$, we simply find the most possible function for $A$, and condition on $A$. It's not hard to see the following:

**Proposition 8** *For any CPM there is an admissible sequence of probability functions such that $P_A^*(B) = P_A(B|A)$. For each admissible sequence, there is a CPM such that the same relation holds.*

Clearly, this definition of admissible sequence relies crucially on the fact that there are a finite number of worlds. However, it suggests an obvious generalization of our CPMs to deal with infinite languages. We simply postulate a sequence of arbitrary indexed probability functions suitable for the language in question. The most possible function in the sequence represents the agent's current epistemic state. Revision by $A$ is simply a matter of finding the most possible function that satisfies $A$, then conditioning by $A$ with respect to that function. In order to mimic the structure of CPMs, we would have to insist that "maximal conjunctions" or possible worlds have positive probability for no more that one function. This is impossible to impose logically since worlds correspond to "infinite conjunctions." It turns out, however, that imposing such a constraint has *no effect on the results of revision*.[6] In essence, allowing worlds to be assigned more than one possibility value (i.e., permitting "duplicate worlds") has no effect on revision, since only the *most* possible value will ever have an influence on our deliberations. Thus, specifying an arbitrary sequence of probability functions, ordered by degree of possibility, is a sound (and very general) representation mechanism for counterfactual probabilities.

There is a second method one might use to reason with counterfactual probabilities using standard probabilistic representations.[7] We can combine the sequence of probability functions into one function if the set of worlds assigned to any degree of possibility is finitely characterizable. This simply means that each cluster of equally possible worlds corresponds to some finite theory, or sentence $\alpha$. In the case of our finite language, this must be true.[8] In such a case, we need only index each possibility value by its characterizing sentence and incorporate this sentence during conditionalization. Formally, we require a sentence $\alpha_k$, for $k > 0$, such that

$$\|\alpha_k\| = \{w : \pi(w) = k\}$$

An arbitrary sequence of sentences is *admissible* iff the elements of the sequence are pairwise disjoint: $\alpha_j \vdash \neg \alpha_k$ iff $j \neq k$. We can assign an arbitrary positive probability weight to each possible world, defining a single probability function $P$. To revise by $A$, we need to find the most possible characterizing sentence consistent with $A$: the $\alpha_k$ such that

$$k = \max\{i : \alpha_i \not\vdash \neg A\}$$

We denote this sentence $\alpha_A$. To revise by $A$, we must condition on $A$; but we are only interested in the most possible $A$-worlds, those that satisfy $\alpha_A$. It's not hard to see the following:

**Proposition 9** *For any CPM there is an admissible sequence of sentences and a probability function $P$ such that $P_A^*(B) = P(B|A \wedge \alpha_A)$. For each admissible sequence and probability function $P$, there is a CPM such that the same relation holds.*

Notice that the single probability function $P$ used to simulate the counterfactual probability function cannot be given a reasonable intuitive interpretation. Indeed, $P$ assigns less than certain probability to full beliefs; and it may make impossible sentences more probable than full beliefs! Function $P$ should be understood as simply a technical device to allow "non-counterfactual" probabilistic reasoning methods to be applied. Thus, one need not define new reasoning mechanisms to deal with counterfactual probabilities.

## 4 Generalized Imaging[9]

Lewis (1976) proposed a method for probabilistic updating known as *imaging* that generally gives results different from those for conditionalization. In a possible worlds framework, the distinction can be understood as follows. We assume a probability function $P$ is determined by an assignment of weight to a finite set of possible worlds. When an epistemic state is updated by sentence $A$ through conditionalization, the weight assigned to $\neg A$-worlds is retracted and the remaining weight (assigned to $A$-worlds) is normalized. This can be thought of as taking the totality of weight $P(\neg A)$ and redistributing it among the $A$-worlds: each $A$-world gets a share of this total in accordance with its relative weight among all $A$-worlds. Notice that conditionalization by $A$ is undefined if $P(A) = 0$, for the relative weight of individual $A$-worlds has no meaning in this context.

---

[5]Using a sequence of probability functions is suggested informally by Lewis (1976).

[6]To see this, imagine that two distinct functions $P_j$ and $P_k$ were such that each assigned positive probability to some possible world (infinite conjunction) $w$. (We assume $j < k$.) The probability assigned to $w$ by the less possible function $P_j$ can influence the result of revision by $A$ only if: a) $P_j$ is the most possible function for $A$; and b) $w \models A$, for the updated function is given by $P_A^*(B) = P_j(B|A)$. But if $w \models A$ then $P_k(A) > 0$ also, contradicting the fact that $P_j$ is most possible for $A$.

[7]Thanks to Fahiem Bacchus for suggesting this representation.

[8]For arbitrary languages and finite conditional *KB*s, this will be the case should one choose, say, a unique most compact model of *KB*. For example, Pearl's (1990) System Z has this property, as does the most compact possibility ranking of (Benferhat, Dubois and Prade 1992). Furthermore, there will be only a finite number of possibility values assigned to possible worlds.

[9]In this section, we present a technical result that is somewhat orthogonal to the rest of the paper.

Lewis's imaging can be thought of as a different way of effecting the minimal change of $P$. Each world $w$ is assumed to have a unique *most similar* $A$-world, denoted $f(w, A)$, that is most like $w$ in relevant respects and satisfies $A$. The function $f$ is a *selection function* that picks out this most similar $A$-world. As with conditionalization, when an update $A$ is to be achieved, weight must be appropriated from $\neg A$-worlds and assigned to $A$-worlds. However, Lewis claims that it is reasonable to expect that the weight taken away from a $\neg A$-world not be arbitrarily distributed among all $A$-worlds. Rather one should assign the weight taken from $w$ to its most similar counterpart satisfying $A$, namely $f(w, A)$. It is clear that typically such a method of update will yield results different from conditionalization. It is also clear that imaging by $A$ is meaningful even when $P(A) = 0$. Notice that one need not distinguish $A$-worlds from $\neg A$-worlds in the redistribution of weight. If we insist that the selection function be *centered*, that is if $f(w, A) = w$ whenever $w \models A$, then we simply redistribute the weight from *every* world $w$ to $f(w, A)$. If $f$ is centered, $A$-worlds (in effect) keep their own weight.

Gärdenfors (1988) describes a slightly more general form of imaging known as *generalized imaging*. This form of update proceeds as with imaging, except that the most similar $A$-world for $w$ need not be unique. Instead, we let $f(w, A)$ denote a *set* of $A$-worlds, those that are most similar to $w$. When imaging by $A$, the weight taken from some $\neg A$-world is redistributed in the appropriate proportions among the worlds in $f(w, A)$. This clearly adheres to the spirit and intent of Lewis's notion. Formally, we have:

**Definition 8** Let $V$ be a (finite) set of worlds and $P$ map $V$ into $[0, 1]$. We let $P$ also denote the probability function (on sentences) induced by this mapping of $W$. A *selection function* is a mapping $f$ from $W \times \mathbf{L}_{CPL}$ into $2^W$.[10] An revision function $*$ is a *generalized imaging* function iff there exists a selection function $f$ such that $P_A^*(w) =$

$$\sum_{v \in W} \{P(v) \cdot \frac{P(w)}{\sum\{P(u) : u \in f(v, A)\}} : w \in f(v, A)\}$$

Our method of update using counterfactual probabilities can be viewed as a generalized imaging function simply

---

[10] Typically we impose restrictions on $f$ (e.g., so that semantically equivalent formulae determine the same most similar worlds), but these are of no concern here.

by taking the set $Pl(A)$ to denote the set $f(w, A)$ for each world $w$. If we wish, we can also use a centered selection function defined as $f(w, A) = \{w\}$ for all $A$-worlds and $f(w, A) = Pl(A)$ for $\neg A$-worlds. With this we see:

**Theorem 10** *Let $M$ be a CPM that induces factual probability function $P$ and revision operator $*$ (that is, the revision of $P$ by $A$ is given by $P_A^*$). Then $*$ is a generalized imaging function.*

Together with Theorem 7 this shows that a generalized imaging function can be constructed in such a way that it accommodates conditionalization. This stands in sharp contrast with the claim of Gärdenfors (1988, Ch.5) that conditionalization and generalized imaging are fundamentally incompatible. Our method of revision using counterfactual probabilities, in fact, embraces both approaches. We explore the full implications of this unification in a longer version of this paper.

## 5 Concluding Remarks

We have presented a semantics for default reasoning and belief revision that admits degrees of belief. Our system incorporates the key aspects of both qualitative and quantitative methods, capturing the statics of representation and the dynamics of revision. It extends both AGM revision and conditionalization, and can be viewed as a form of generalized imaging. It can also be seen as a unification of probability and possibility theory.

One issue that remains unaddressed in this paper is the derivation of new counterfactual probabilities after update, and how to iterate the process of revision. The updated function $P_A^*$ is specified only for factual probabilities, and determines the new *objective* epistemic state. Counterfactual conditional probabilities $P_A^*(C \uparrow B)$ remain unspecified when $P_A^*(B) = 0$. So imagine one updates $P$ by $A$. If $P_A^*(B) = 0$, then the result of updating $P_A^*$ by $B$ is cannot be determined. There are several directions in which one might proceed. One method extends the qualitative *natural revision* model of (Boutilier 1993c). In this approach, the worlds in $Pl(A)$ become most possible while all other worlds retain the same (relative) ranking of possibility. If the relative probability weight of each world remains unchanged, this model provides a method of updating counterfactual probabilities in which as few counterfactual probability values as possible are altered. A somewhat different mechanism would adopt the method of *J-conditioning* proposed by Goldszmidt and Pearl (1992) for updating conditional rankings. However, such a method involves the updating the possibility distribution so that worlds that originally had different degrees of possibility now can have the same degree. How to reconcile the relative weights of such worlds in this case is not clear. Related to these proposals is the model of arbitrary conditional revision proposed in (Boutilier and Goldszmidt 1993). Such a model, if extended, would allow one to update counterfactual probabilities without changing factual probabilities.

Another issue that remains unaddressed is how factual prob-abilities can influence or determine counterfactual probabilities. This is a difficult problem, even in qualitative belief revision. The impact of the AGM theory (and our model here) is to suggest that, in general, one must permit arbitrary changes. Practically speaking, this is unsatisfying, and practical constraints on updating probabilities must be investigated. For example, if $A$ is "unrelated" to $B$ and $C$, and $P$ is updated by $A$, we would like the conditional probability $P(C|B)$ to influence (perhaps determine) the new probability $P_A^*(C|B)$, even if $P(A) = 0$.

We are currently exploring the use of counterfactual probabilities in model-based diagnosis. In (Boutilier and Becher 1993), we have embedded logical approaches to diagnosis in the qualitative conditional framework. Conditional defaults allow one to completely (but defeasibly) discount potential candidate diagnoses. Attaching (counterfactual) probabilities to such rules allows one to represent the fact that certain of the *remaining* candidates are more likely than others. This provides a semantics that might underly methods of logical diagnosis that incorporate probabilistic information (de Kleer 1991; Poole 1993).

## Acknowledgements

I would like to thank Fahiem Bacchus, Moisés Goldszmidt, Keiji Kanazawa and David Poole for their helpful comments and suggestions. This research was supported by NSERC Research Grant OGP0121843.